\documentclass[10pt,letterpaper]{article}

\usepackage{cogsci}

\cogscifinalcopy 

\usepackage[pdftex]{graphicx}
\usepackage{float}
\usepackage{url}
\usepackage[percent]{overpic}
\usepackage[
  style=apa,
  natbib=true,
  annotation=false,
]{biblatex}
\usepackage{float} 

\title{Large Language Models Can Take False First Steps at Inference-time Planning}

\author[1]{\mbox{Haijiang Yan (haijiang.yan@warwick.ac.uk)}}
\author[2]{\mbox{Jian-Qiao Zhu}}
\author[1]{\mbox{Adam Sanborn}}
\affil[1]{Department of Psychology, The University of Warwick}
\affil[2]{Department of Psychology, The University of Hong Kong}

\begin{document}

\maketitle

\begin{abstract}
Large language models (LLMs) have been shown to acquire sequence‑level planning abilities during training, yet their planning behavior exhibited at inference time often appears short‑sighted and inconsistent with these capabilities. We propose a Bayesian account for this gap by grounding planning behavior in the evolving generative context: given the subtle differences between natural language and the language internalized by LLMs, accumulated self-generated context drives a planning-shift during inference and thereby creates the appearance of compromised planning behavior. We further validate the proposed model through two controlled experiments: a random‑generation task demonstrating constrained planning under human prompts and increasing planning strength as self‑generated context accumulates, and a Gaussian‑sampling task showing reduced initial bias when conditioning on self‑generated sequences. These findings provide a theoretical explanation along with empirical evidence for characterizing how LLMs plan ahead during inference. 

\end{abstract}

\section{Introduction}

Despite the fact that LLMs generate text by predicting the next token auto‑regressively, a growing body of evidence indicates that they engage in sequence‑level planning \citep{blondel2025autoreg, pal2023future, men2024unlocking, pochinkov2024extracting, wu2024language, lindsey2025biology, cencerrado2025no, dong2025emergent}, rather than operating in a purely “short‑sighted’’ manner. \citet{blondel2025autoreg} establish a theoretical equivalence between next‑token prediction in LLMs and sequence‑level modeling in energy‑based models, demonstrating that training LLMs for auto‑regressive prediction is, in effect, training them to plan at the sequence level. Under this view, an LLM should maintain awareness of the full set of future continuations it is implicitly planning for before any individual token is generated. Empirically, however, related evidence for such planning remains limited: studies instead report short‑term planning patterns during inference \citep{pal2023future, men2024unlocking, cencerrado2025no}, suggesting a gap between the systematic planning abilities LLMs acquire through training and the behaviors they actually exhibit at inference time. At present, the mechanism underlying this gap remains poorly understood.

In principle, next‑token training should equivalently induce sequence‑level planning \citep{blondel2025autoreg}. Given an prompt sequence $I_{\mathcal{C}}$ sampled from training corpus $\mathcal{C}$, the pretraining objective that determines the probability assigned to next tokens can be expressed probabilistically as:
\begin{equation}
\label{full_planning}
    P(x|I_{\mathcal{C}}) = \int_s P(x|s)P(s|I_{\mathcal{C}}) ds
\end{equation}
where $s$ ranges over all possible response sequences consistent with $I_{\mathcal{C}}$. LLMs are thereby encouraged to consider the full set of candidate sequences $s$ before pinning down the distribution over next token $x$. This perspective aligns with empirical findings showing that future information is more accurately decoded from early‑ and mid‑layer activations, whereas accuracy on the immediate next token peaks only at the final layer \citep{dong2025emergent, pal2023future, cencerrado2025no}. Notably, \citet{jenner2024evidence} reported a similar ``plan-first'' pattern in a chess-playing neural network, suggesting that such planning behavior may arise generally in models trained on sequential data.

However, inference‑time behavior of LLMs appears to be more complex. Empirical work that decodes future tokens from early neural activations suggests that their planning capacities primarily govern short‑term future generations rather than extending uniformly across the whole response \citep{pal2023future, men2024unlocking}. This raises a central question we aim to answer in this paper: if LLMs are trained on $\mathcal{C}$ to perform sequence‑level planning as expressed in Equation \ref{full_planning}, what constrains their ability to fully plan ahead at inference.

A key distinction at inference time lies in the inherent mismatch between the distribution of the training corpus $\mathcal{C}$ (i.e., natural language produced by humans) and LLM-generated data $\mathcal{M}$ (what LLMs actually represent). Prior work has documented such discrepancies both in specific tasks \citep{harrison2024comparison} and in broader text‑generation settings \citep{an-etal-2024-capturing, zhu2024synthesize}. These findings suggest that, even trained extensively on human data, the generative process of LLMs introduces systematic deviations from the training distribution, yielding stylized outputs that enable a high rate of self-recognition \citep{panickssery2024llm}. Consequently, an arbitrary human‑written prompt $I_{\mathcal{C}}$ may constitute an out‑of‑distribution input for the model at inference time. Yet whether—and in what ways—such out‑of‑distribution prompts compromise planning behavior remains unknown. 

With this in mind, our aim in this paper is to reconcile LLMs’ theoretical planning abilities with their actually planning behavior at inference time, with particular attention to how this behavior is shaped by the dynamically evolving context. In the remainder of the paper, we first introduce a Bayesian perspective on sequence‑level planning in LLMs, clarifying how such planning is deployed at inference in ways that can lead to compromised planning behavior. Building on this framework, we derive previously unreported dynamic patterns of planning and then test these predictions empirically. In Experiment 1, using a random‑generation task, we confirm the presence of constrained planning under human‑written prompts and validate the prediction that planning shifts and converges as in‑distribution self‑generated text accumulates during inference. In Experiment 2, we employ a Gaussian‑sampling task with ground‑truth structure to further verify how prompts dynamically shape the evolution of the model’s planning. Overall, our work provides both theoretical perspective and empirical evidence that deepen our understanding of how LLMs plan at inference.

\section{A Bayesian Account of Sequence-level Planning}

\subsection{Meta-learned Bayesian Prior}
LLMs are generally known as excellent few-shot \citep{NEURIPS2020_1457c0d6} or even zero-shot learners \citep{kojima2023largelanguagemodelszeroshot, wei2022finetunedlanguagemodelszeroshot}. That is, they can solve a variety of tasks by simply conditioning on a few task examples or merely instructions of the task. Meta-learning over prior across tasks provides a principled explanation for such quick adaption to sparse or no new data \citep{chen2022metalearninglanguagemodelincontext, finn2019probabilisticmodelagnosticmetalearning, kirsch2024generalpurposeincontextlearningmetalearning}. Suppose a domain-specific generation task distribution $\phi=\{\tau_1, \tau_2, \dots \tau_n\}$ is embedded in the training data. Combined with the training objective in Equation \ref{full_planning}, an LLM trained to capture such a domain‑specific planning prior can be described as
\begin{equation}
    P(s|\phi) = \int_\tau P(s|\tau, \phi) d\tau
\end{equation}

The priors encoded by LLMs have been shown to align closely with those employed by humans \citep{jagadish_meta-learning_2025}. Given such priors, Bayesian inference over response sequences becomes possible, enabling LLMs to perform few‑shot and even zero‑shot learning. At inference time given a task prompt $I$ associated with domain $\phi$, the model infers a response sequence $s$ by combining the encoded domain prior with the prompt information in a Bayesian manner:
\begin{equation}
\label{inference_bayes}
  P(s|I, \phi) \propto 
  \underbrace{P(I|s, \phi)}_{\text{planning likelihood}} \; \cdot
  \underbrace{P(s|\phi)}_{\text{domain prior}}
\end{equation}
The planning term $P(I|s, \phi)$ measures how well the prompt $I$ is explained by a planned sequence $s$ under domain $\phi$, whereas the domain prior $P(s|\phi)$ reflects the distribution over sequences induced solely by the domain‑level structure learned during training.

\begin{figure*}[t!]
    \begin{center}
        \includegraphics[width=\linewidth, trim={0cm 4cm 0cm 4cm}, clip]{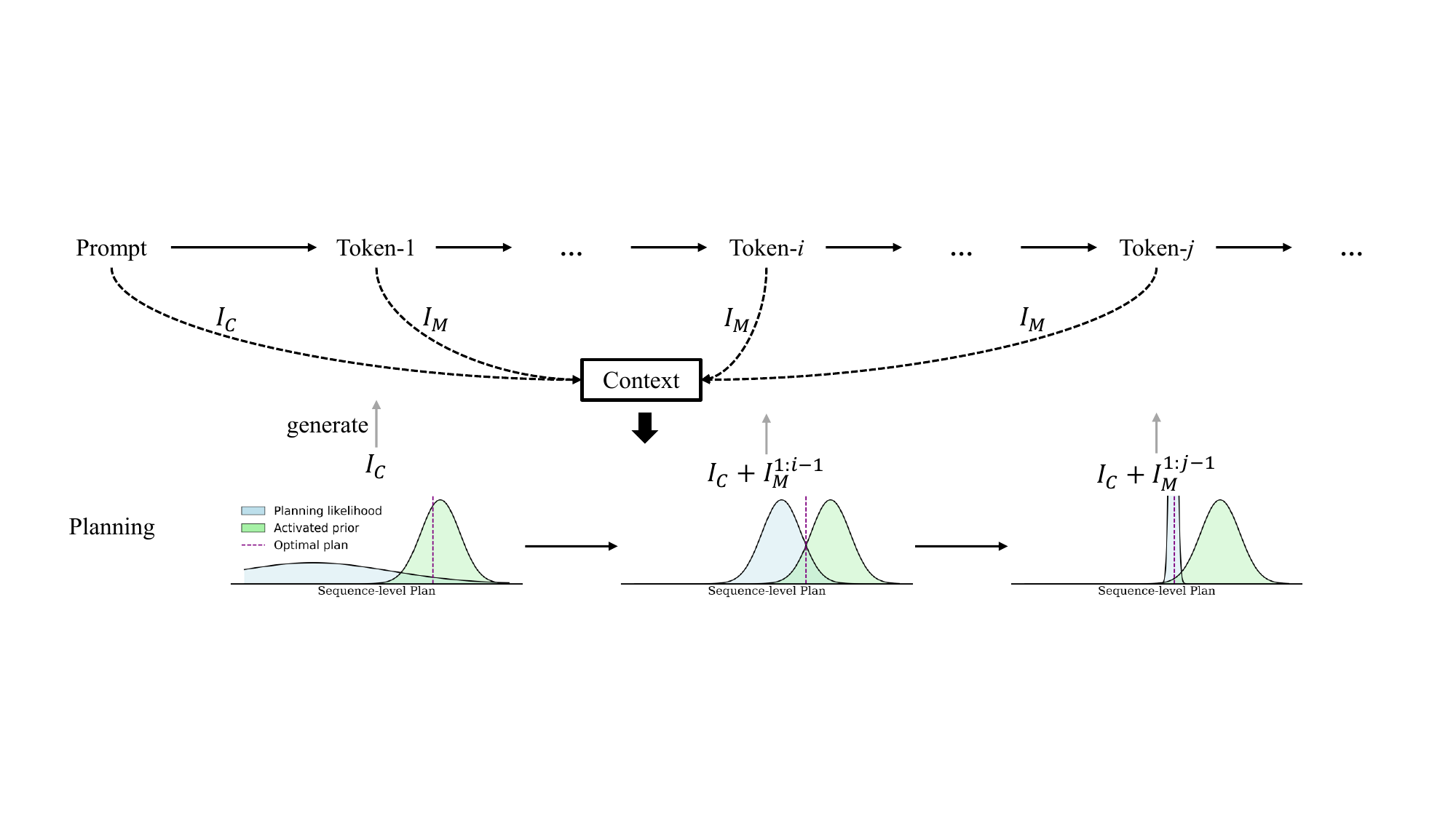}
    \end{center}
    \caption{Scheme illustrating the evolving planning behavior of LLMs during inference.}
    \label{process}
\end{figure*}

\subsection{Compromised Planning during Inference}
Inherent differences between natural‑language prompts $I_{\mathcal{C}}$ and truly internalized language $I_{\mathcal{M}}$ may underlies LLMs' compromised planning observed at inference time through Equation~\ref{inference_bayes}. Ideally, an LLM trained on human dialogues should assign high planning likelihood to good response $s$, enabling instruction‑following behavior and coherent multi‑sentence responses. Thus, for an in‑distribution prompt $I_{\mathcal{M}}$, the planning likelihood concentrates with high certainty on sensible continuations $s$ drawn from the model’s internalized corpus $\mathcal{M}$. In such cases, planning behavior is dominated by the planning‑likelihood term, and the prior exerts minimal influence on the resulting token generation. As generation unfolds, the model’s own self‑generated tokens make the context increasingly resemble its represented distribution, causing the planning likelihood to sharpen further and reinforcing the initial planning trajectory.

However, because LLMs represent $I_{\mathcal{C}}$ and $I_{\mathcal{M}}$ in inherently different ways, conditioning on $I_{\mathcal{C}}$ introduces substantially greater uncertainty into the planning‑likelihood term at inference time: the set of candidate continuations $s$ that count as reasonable responses to $I_{\mathcal{C}}$ becomes broader:
\begin{equation}
    \mathcal{H}[P(s|I_{\mathcal{C}}, \phi)] > \mathcal{H}[P(s|I_{\mathcal{M}}, \phi)]
\end{equation}
where $\mathcal{H}$ represents an entropy over candidate responses. In such cases, the prior term exerts greater influence on sequence planning. 

The prior‑driven planning ensures that the model can produce a reasonable response to $I_{\mathcal{C}}$ at the beginning of inference under high uncertainty, yet it can also bias the model away from faithfully following $I_{\mathcal{C}}$---for example, producing an answer that looks acceptable but actually inaccurate. Importantly, this does not imply that the planning mechanism itself is impaired; the model can still adhere to whatever plan is inferred at first. 

However, the evolving context drives shift of planning. As the self-generated text (effectively functions as new $I_{\mathcal{M}}$) is concatenated back into $I_{\mathcal{C}}$, the uncertainty in the planning‑likelihood term gradually decreases over the course of generation, allowing the likelihood aligned with the instruction to regain dominance (as illustrated in Figure~\ref{process}). In this ``prior-suppressing'' process, the first observable consequence at inference time is the planning shift, in which sequence‑level planning shifts from prior‑preferred continuations to likelihood‑preferred ones. As a result, a plan (embodied in hidden activations) occurred early during generation cannot reliably predict later tokens, giving rise to the short‑term planning pattern observed empirically.

\subsection{Predicted Dynamic Signatures of Planning}
The dynamic planning process depicted above can be summarized as:
\begin{equation}
    P(s_t|I^{(t)}_{\mathcal{C}\rightarrow \mathcal{M}}, \phi) \propto P(I^{(t)}_{\mathcal{C}\rightarrow \mathcal{M}}|s_t, \phi) \; \cdot P(s_t|\phi), \quad t=1, 2\dots T
\end{equation}
The changing context continually adjusts the position as well as extent of the likelihood, forcing LLMs to adapt the internal planning behavior to context to assure an optimal plan at each inference step. 

Importantly, this shift does not continue indefinitely; instead, it will gradually slow and converge toward a likelihood-preferred plan as in-distribution $I_{\mathcal{M}}$ accumulates in the context. Consequently, a characteristic signature predicted by the model is that the compromised planning behavior progressively recovers as planning stops shifting and converges to an optimal plan over the course of inference.

Beyond the \textit{planning shift and convergence}, this process model implies another central consequence: planning begins from a prior‑biased state and gradually converges toward an unbiased plan that faithfully responds to the query. More specifically, because the shifting plan must continually adjust to the context in order to realize an unbiased answer, the generation process will naturally exhibit a \textit{biasing-then-debiasing} dynamic. 

Overall, by grounding how LLMs generalize their acquired planning ability to new data (e.g., a novel user query outside the support of training data) under uncertainty, the planning model proposed in this paper offers a fresh perspective on the compromised planning behavior observed at inference time. By assuming only that LLMs experience higher uncertainty when interpreting natural language, the model seamlessly captures the short‑term planning pattern as a consequence of planning shift. In addition, the model yields further testable predictions about planning dynamics, which we validate through two experiments described below.

\section{Experiment 1: Planning Shift and Convergence in Random Generation Task}
\label{Experiment1}

If accumulated in‑distribution self‑generated text (denoted $I_{\mathcal{M}}$) drives \textit{planning shift and convergence}, LLMs should exhibit an increase in planning strength over the course of generation. To verify such dynamics of planning behavior in sequential token generation, we adopt a random number generation task from \citet{castillo2024explaining}. The task was chosen for its simplicity, as it produces outputs in a one-dimensional numeric space where predictive relationships can be easily measured. 
Using open-source LLMs (Llama-3.1-8B-Instruct and Qwen-2.5-7B-Instruct), we elicited random numeric estimates of human height to probe how models form and commit to future answers in advance. What makes this task appealing for evaluating plans is that the models are prompted to produce independent random estimations consecutively, thereby minimizing confounds from auto-correlated continuations within sequences that LLMs may produce. The prompt used is as follows:
\begin{quote}
    \textbf{Prompt.}
    ``You are simulating guesses of adult heights (in centimeters) for individuals randomly drawn from the UK population. In each round, output a single integer as your guess for the random person’s height. Continue producing guesses one after another, separated by commas, with no explanations or extra text: \textit{\{starting point\}}, ''
\end{quote}

The temperature of generation is set at 0 (greedy decoding) to maximize the planning behavior. In each trial, LLMs generated 60 estimates in sequence, with each pair of consecutive outputs separated by a comma and a space (e.g., ``182, 163, 159, ...''). Each model completed 69 trials, each initialized with a unique starting value ranging from 151 cm to 220 cm. Thus, every model produced 69 distinct sequences of height estimates, with each sequence beginning with a number between 151 and 220 and containing a total of 60 values. Beside the completions, the embeddings (with dimensions of $d=3584$ for the Qwen model and $d=4096$ for the Llama model) for all new tokens from both models were also collected to reflect the information on which each new estimate was built.

\begin{figure}[t!]
  \begin{center}
    \includegraphics[width=\linewidth]{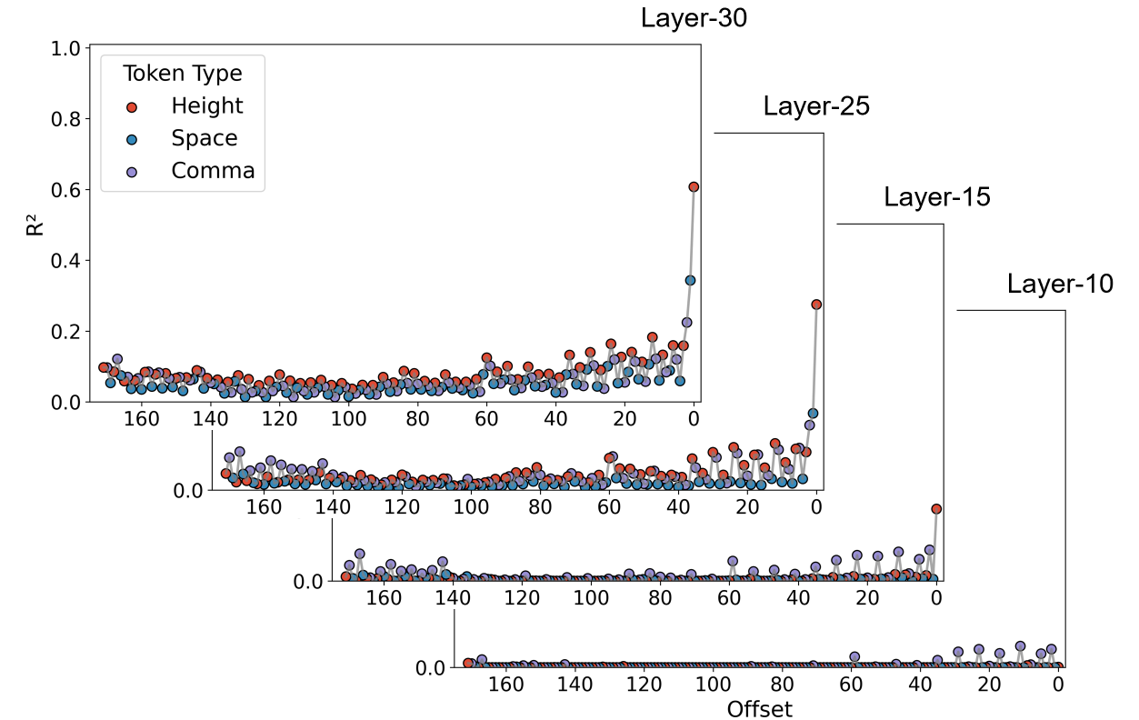}
    \includegraphics[width=\linewidth]{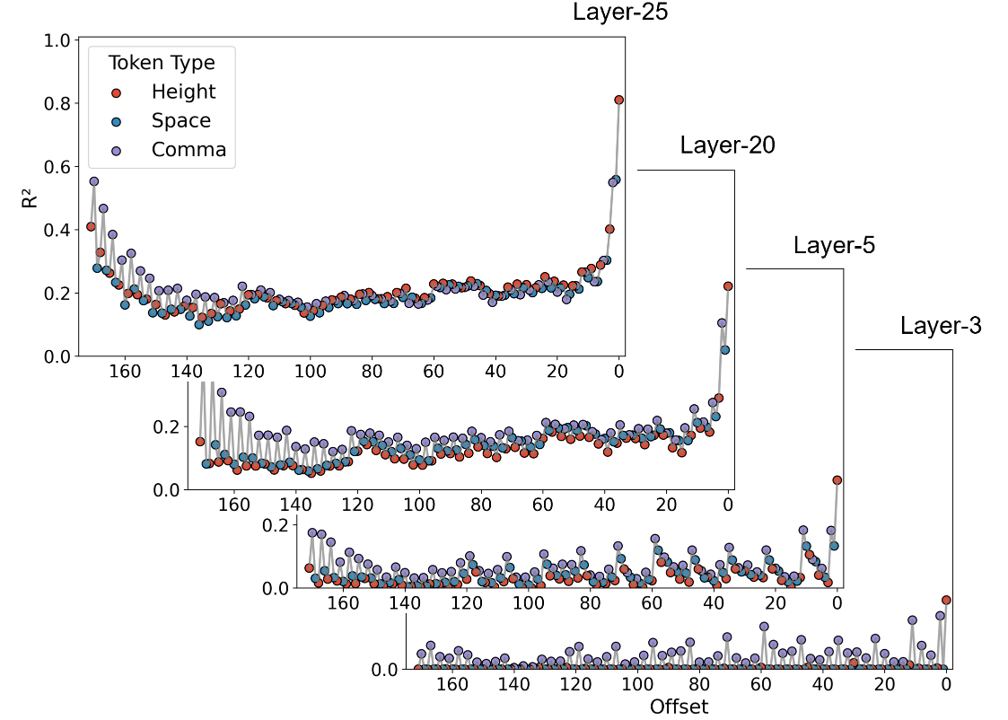}
  \end{center}
  \caption{Percentage of variance in future tokens ($R^2$ of regressions; vertical axis) captured by the LLM embeddings across different planning horizons (offset $\Delta t$; horizontal axis): \textbf{(top)} Llama-3.1-8B-Instruct, \textbf{(bottom)} Qwen-2.5-7B-Instruct.}
  \label{regression_results}
\end{figure}

\subsection{Data Analysis}
We evaluate how well the LLM embedding extracted at time $t$, $\phi_t$, predicts future tokens $y_{t+\Delta t}$ with a look-ahead offset of $\Delta t$. This analysis has two implications. First, if $\phi_t$ is predictive of $y_{t+\Delta t}$, it suggests that the LLM engages in forward planning. Second, we are also interested in how the offset $\Delta t$ modulates the predictive relationship. If smaller offsets yield stronger predictability, this would suggest a short‑term planning pattern as suggested by prior research.

We applied LASSO regression with a penalty coefficient $\alpha$ as 0.3 to estimate the prediction performance of embeddings $\mathcal{\phi}_{t}=f(x,y_{<t})\in\mathbb{R}^{n\times d}$ in forecasting a future response value $y_{t+\Delta t}\in\mathbb{R}^{n}$ ($\Delta t\geq 0$). The offset parameter $\Delta t$ ranged from 0 to 172, reflecting the lag between embeddings and subsequent numeric outputs (corresponding to a prediction horizon of 0–58 numeric samples, as commas and spaces were included in the tokenization). For each offset, all available training data $\{\mathcal{\phi}_{t}^{(n, :)}, y_{t+\Delta t}^{(n, :)}\}_{n=1}^{69}$ from 69 sequences were extracted and aggregated, and a separate regression model was fitted to evaluate the degree to which embeddings at that offset encode information about upcoming responses.
\begin{equation}
\label{lasso1}
    y_{t+\Delta t}=\mathcal{\phi}_{t} \omega_{\Delta t} + b_{\Delta t}, \; 0\leq \Delta t\leq 172
\end{equation}
The regression coefficients $\omega_{\Delta t}\in\mathbb{R}^{d}$ on future values then could represent the planning behavior at given positions. The corresponding fitting goodness, $R^2$, indicates the strength of the planning across specific planning horizon. 

Then, to further examine how planning strength evolves as $I_{\mathcal{M}}$ accumulates over the course of generation, we fixed the planning horizon at $\Delta t = 8$, which provides a fair window for quantifying predictive planning. At each comma position $q$, we constructed datasets $\{\mathcal{\phi}_{q}^{(n)}, y_{q+8}^{(n)}\}_{n=1}^{69}$ and fitted a separate LASSO regression model. This allowed us to track how well embeddings eight steps beforehand anticipate responses at position $q$, thereby revealing the dynamics of planning signals across the sequence.
\begin{equation}
    y_{q+8}=\mathcal{\phi}_{q} \omega_{q} + b_{q}, \; q=3i+1, \; 0\leq i\leq 57
\end{equation}

\subsection{Results}

In the random generation task, positive percentages of variance in future tokens captured by the embeddings suggest the existence of planning behavior across almost all layers of both LLMs (see Figure~\ref{regression_results}). In particular, both LLMs begin planning for future token generations from the early layers at the comma positions following each sample. At the sample positions themselves, the models first reflect on the current token in the early to middle layers before planning ahead in the later layers. Importantly, the strength of planning decreases drastically as the offset increases, confirming the short-term planning pattern at inference.

\begin{figure}[t!]
    \begin{center}
        \includegraphics[width=0.49\linewidth]{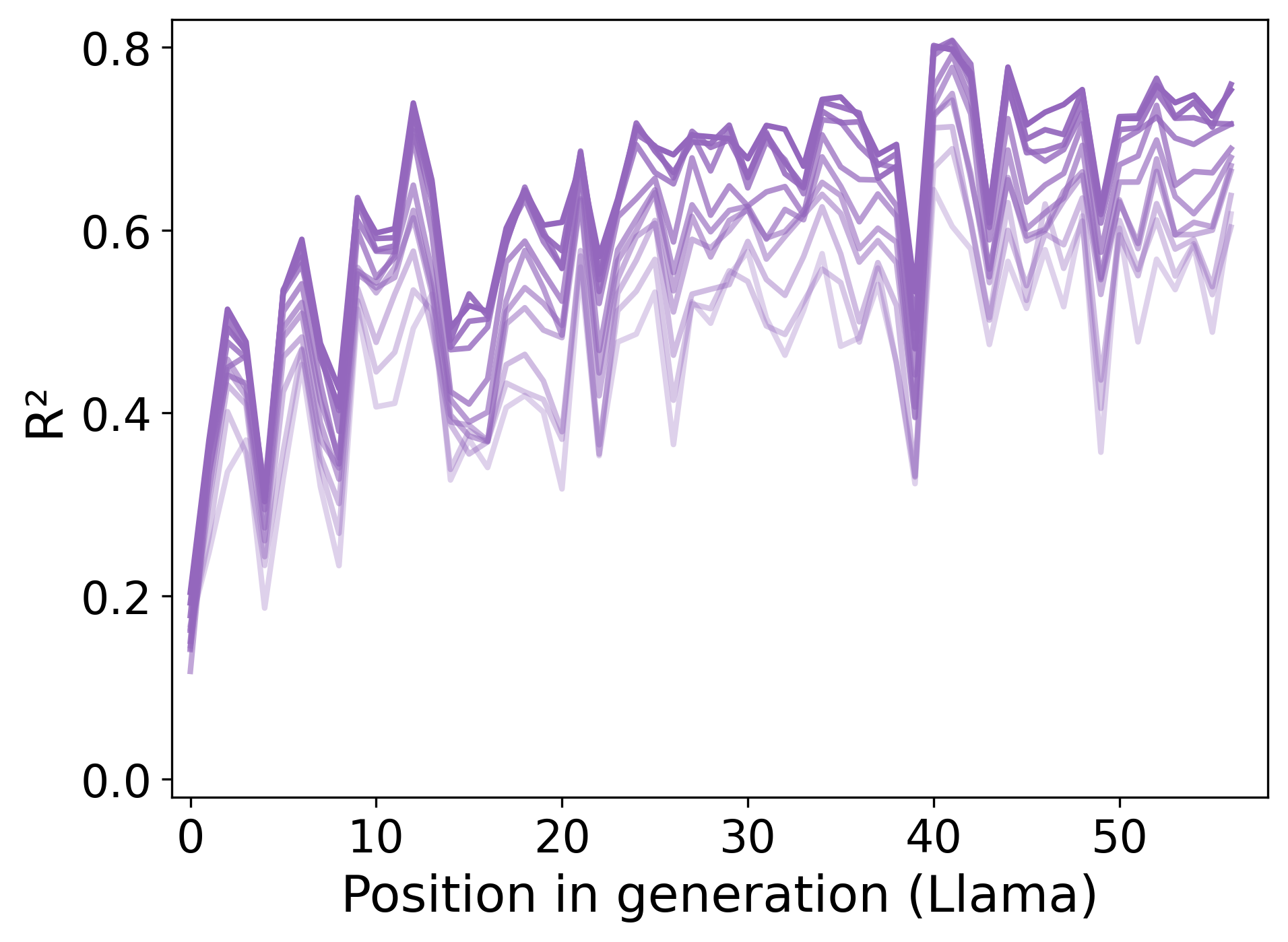}
        \includegraphics[width=0.49\linewidth]{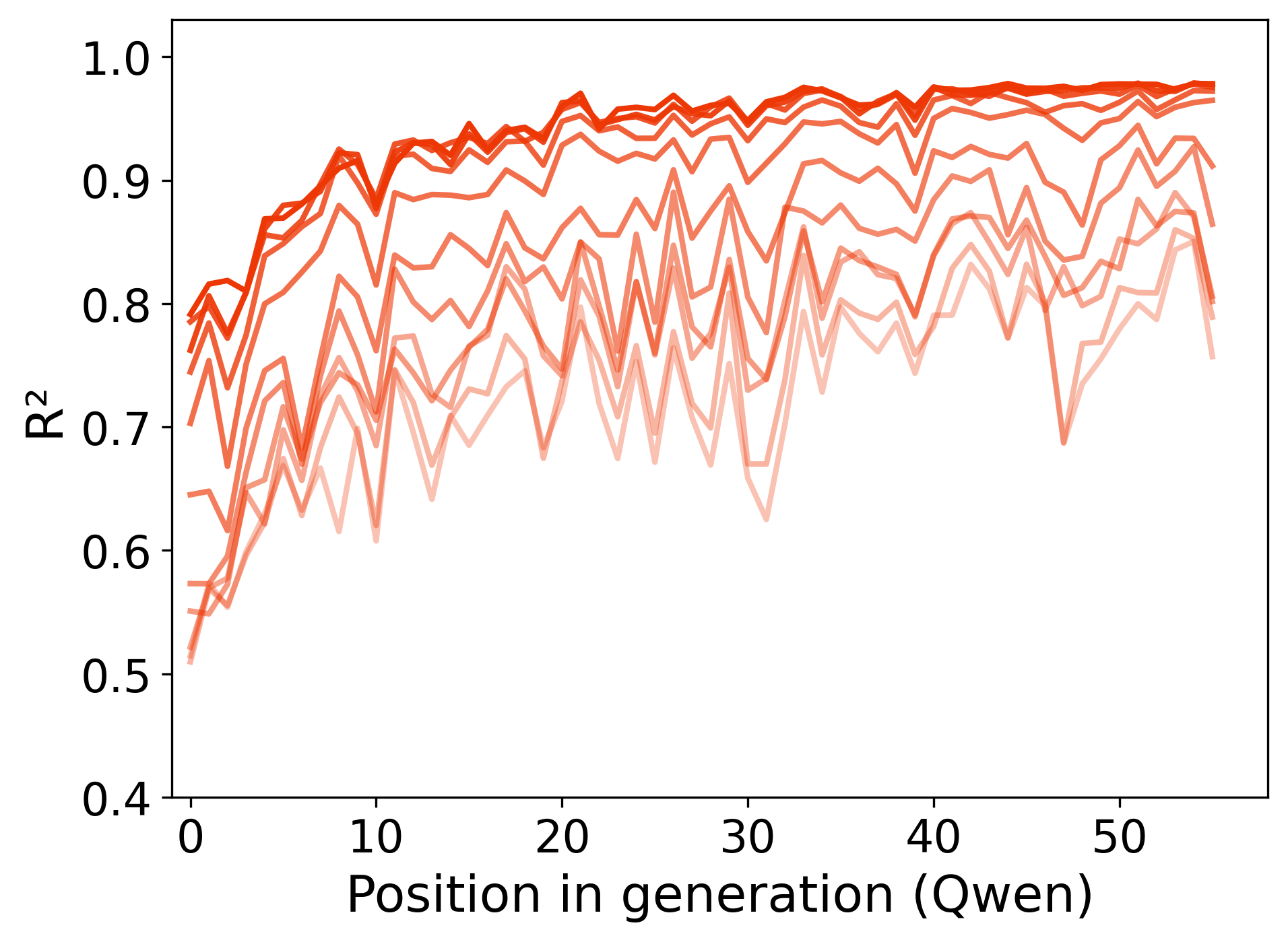}
    \end{center}
    \caption{Relationships between plan adherence and planning position: \textbf{(left)} Llama-3.1-8B-Instruct, layer 15-25 from lightest to darkest, \textbf{(right)} Qwen-2.5-7B-Instruct, layer 10-20 from lightest to darkest. Higher $R^2$ indicates stronger adherence of later token generations to the latent plans formed eight steps earlier.}
    \label{convergence}
\end{figure}

However, the planning strength increases, with the $R^2$ for predicting future tokens rising to nearly 0.8 for Llama and approaching even 1.0 for Qwen, as more $I_{\mathcal{M}}$ accumulates in the context (Figure~\ref{convergence}). The changing context enables the model to increasingly adhere to its inferred plan after the initial shifting phase, reflecting not only the model’s inherent planning ability but also a converging process.

\subsection{Discussion}

The results in Experiment 1 confirm the \textit{planning shift and convergence} as foreseen by the proposed model, which leads to the short-sighted planning behavior as observed both in previous study and in this experiment. This experiment further highlights the role of $I_{\mathcal{C}}\ vs.\ I_{\mathcal{M}}$ in shaping the dynamic process of planning behavior.

\section{Experiment 2: Biasing-then-debiasing in Gaussian Sampling Task}

Because Experiment 1 lacks ground‑truth labels, it does not allow us to determine whether the observed responses are biased. Experiment 2 therefore extends the analysis by evaluating whether LLMs display the predicted biasing-then-debiasing dynamics under conditions where biases can be explicitly measured.

In Experiment 2, LLMs were prompted to generate random numbers--this time sampled from Gaussian distributions, $\mathcal{N}(\mu, 10)$. We first generated 64 ground-truth samples from a real Gaussian distribution and then prompted the LLM to continue the sequence with an additional 64 samples (i.e., Gen I samples in Figure \ref{generation_dynamics}). Note that the LLMs were not provided with the information about the Gaussian distribution (see prompts below). Next, we conditioned the LLM only on the 64 self-generated samples from Gen I as example samples and again prompted it to continue generating random numbers (i.e., Gen II samples in Figure \ref{generation_dynamics}). This design yielded two comparison groups: one beginning with 64 random samples from the Gaussian, and the other beginning with 64 samples generated by the LLM. 
\begin{quote}
    \textbf{Prompt.}
    ``You are sampling integers from a distribution. You will see some samples that has already been drawn from this distribution and your task is to continue sampling integers from this distribution, separated by commas, with no explanations or extra text. Please continue the sampling: \textit{\{23, 45, 12, 43, 55, 4, \dots, (example samples)\}}''
\end{quote}

\subsection{Data Analysis} 
For this task, the rational plan $\rho$ is to generate numbers randomly from $\mathcal{N}(\mu, 10)$, although the LLMs must infer the Gaussian distribution from the observed context samples. Seven different $\mu$ values, ranging from $-50$ to $50$, were tested in the experiment. Each of the seven conditions was repeated 100 times, with a different set of 64 randomly generated samples prepended in each run. By examining the distributions of LLM-generated outputs at each step within each condition, we compared them against the ground truth (the true conditional distribution, shown as starting squares in Figure~\ref{generation_dynamics}) and recovered the dynamics of output quality over the course of generation. 

\subsection{Results}
\begin{figure*}[t!]
    \begin{center}
        \includegraphics[width=0.45\linewidth, trim={2cm 0cm 0cm 0cm}, clip]{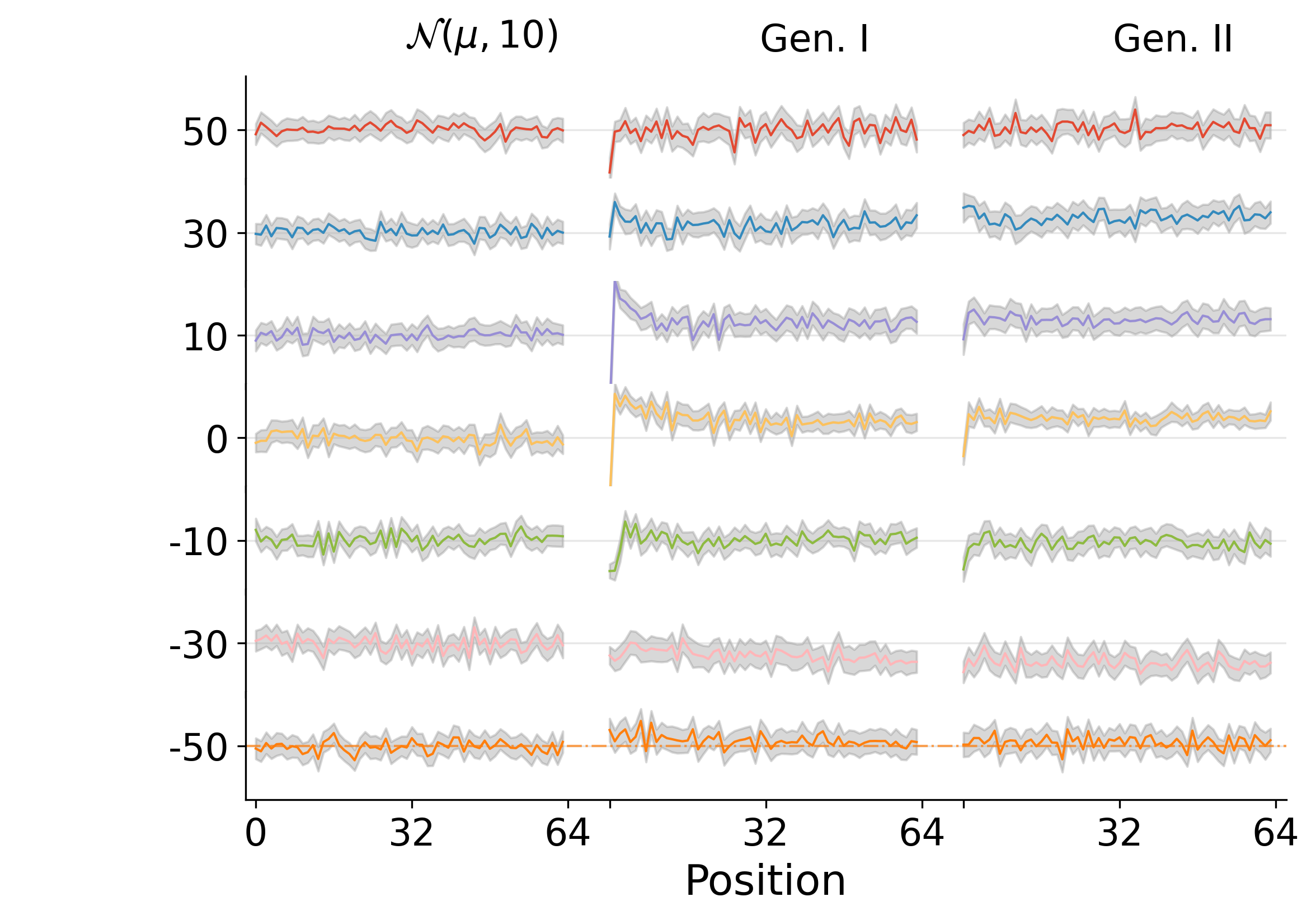}
        \includegraphics[width=0.45\linewidth, trim={2cm 0cm 0cm 0cm}, clip]{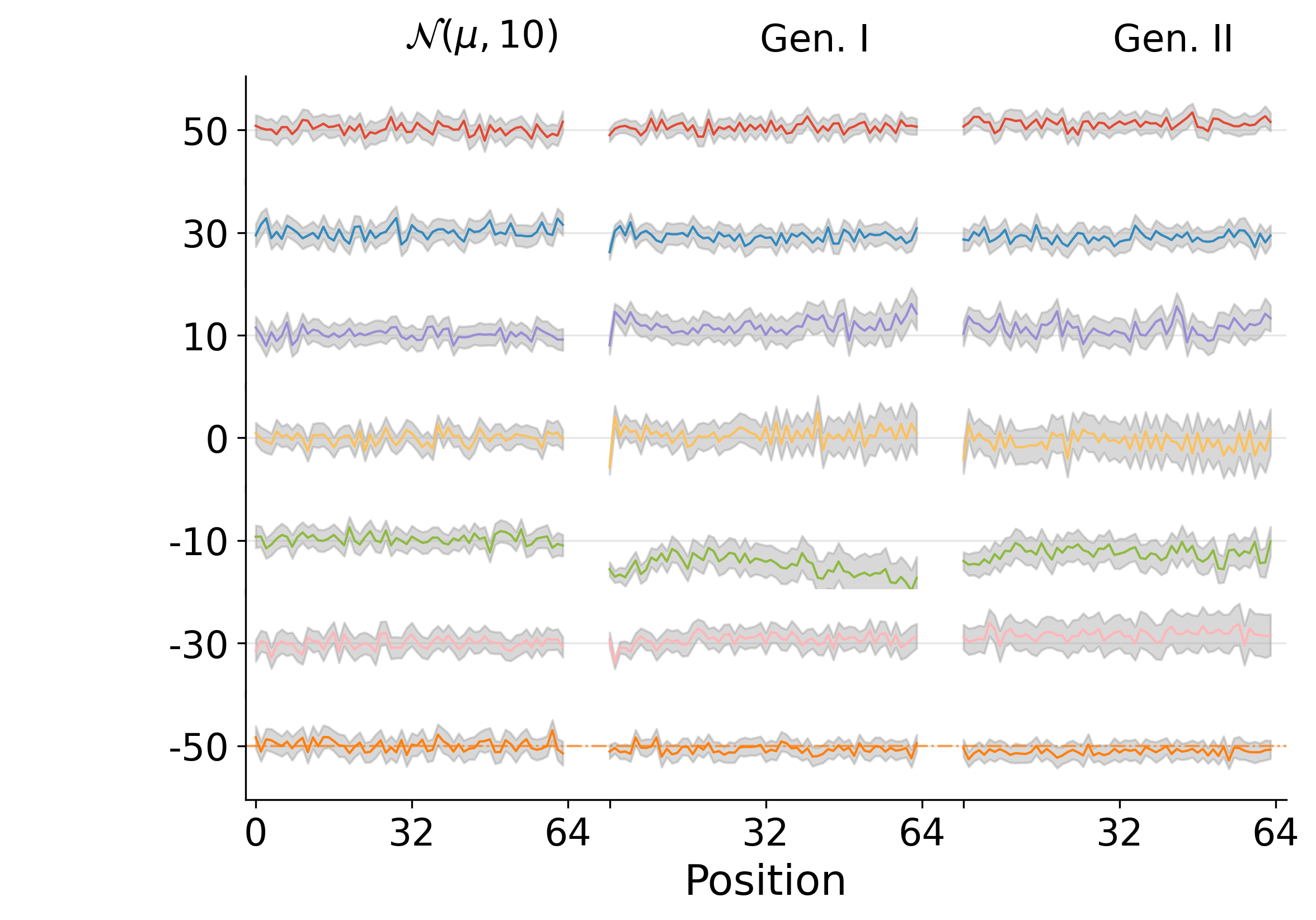}
    \end{center}
    \caption{Generation start-off dynamic patterns in Gaussian sampling task: \textbf{(left)} Llama-3.1-8B-Instruct, \textbf{(right)} Qwen-2.5-7B-Instruct. Both panels display the mean with 95\% confidence intervals of samples at each generation position, with the seven conditions presented vertically. Within each condition, three sets of samples are shown from left to right: (i) samples from a Gaussian distribution $\mathcal{N}(\mu, 10)$, (ii) samples generated by the LLM conditioned on the Gaussian (Gen. I), and (iii) samples generated by the LLM conditioned on Gen. I (Gen. II). We find statistically significant negative biases in the initial samples of Gen. I (see Tables \ref{ttest_starting_bias_llama} and \ref{ttest_starting_bias_qwen} for details).
    }
    \label{generation_dynamics}
\end{figure*}

As shown in Figure~\ref{generation_dynamics}, consistent starting patterns emerged across the seven conditions. For both the Llama and Qwen models, the initial generations were biased toward negative values (around -30) before being de-biased --- sometimes even over-corrected --- eventually converging to stable distributions that closely match the mean of the underlying Gaussian. This \textit{biasing–then–debiasing} pattern aligns with the prediction of the Bayesian model when the initial random samples $y_{\leq 64}$ were not produced by the LLM. When conditioned on self-generated samples, the subsequent generations (i.e., Gen II samples) exhibited less bias compared to Gen I samples, which were conditioned on externally provided random samples from the Gaussian distribution. 

We further quantify the starting bias by comparing samples produced at the first position between the sequences from Gen. I (conditioned on Gaussian) and Gen. II (conditioned on self-generated samples) in each $\mu$ condition. As shown in Table~\ref{ttest_starting_bias_llama} and ~\ref{ttest_starting_bias_qwen}, when conditioned on Gaussian, the LLM presented significant starting bias under more conditions than conditioning on self-generated samples, supporting the prediction that conditioned on self-generated context is able to reduce the starting bias induced by out-of-distribution context. 

\begin{table}[t!]
  \begin{center}
    \caption{Comparison of the first generated samples from Llama-3.1-8B-Instruct}
    \label{ttest_starting_bias_llama}
    \vskip 0.12in
    \begin{tabular}{llll}
      \hline
      Conditions        & Gen. I      & Gen. II      & I vs. II      \\
      \hline
        $\mu = 50$ &43.70 (1.12)** &49.24 (.93) &t=-3.78** \\ 
        $\mu = 30$ &29.41 (.96) &33.68 (1.10)* &t=-2.94* \\
        $\mu = 10$ &-1.91 (1.04)** &9.39 (1.16) &t=-7.26** \\
        $\mu = 0$ &-9.63 (.53)** &-2.71 (.64)** &t=-8.37** \\
        $\mu = -10$ &-14.53 (.53)** &-14.28 (.91)** &t=-.25 \\
        $\mu = -30$ &-31.84 (.67)* &-34.29 (.81)** &t=2.35* \\
        $\mu = -50$ &-47.67 (.87)* &-49.84 (.92) &t=1.72 \\
      \hline
    \end{tabular}
  \end{center}
      \footnotesize
      \textit{Note.} Standard errors are shown in parentheses. 
      Under Gen. I and Gen. II, $*$ represents statistical significance in the comparisons between corresponding generation and the ground-truth $\mu$. 
      $^{*}p < .05$, $^{**}p < .001$. These also apply to Table~\ref{ttest_starting_bias_qwen}.
\end{table}

\begin{table}[t!]
  \begin{center}
    \caption{Comparison of the first generated samples from Qwen-2.5-7B-Instruct}
    \label{ttest_starting_bias_qwen}
    \vskip 0.12in
    \begin{tabular}{llll}
      \hline
      Conditions        & Gen. I      & Gen. II      & I vs. II      \\
      \hline
        $\mu = 50$ &49.23 (.49) &50.50 (.62) &t=-1.63 \\ 
        $\mu = 30$ &27.10 (.56)** &28.98 (.88) &t=-1.87 \\
        $\mu = 10$ &8.51 (.69)* &10.21 (.92) &t=-1.50 \\
        $\mu = 0$ &-4.35 (.54)** &-3.24 (1.04)* &t=-.98 \\
        $\mu = -10$ &-14.26 (.53)** &-13.04 (.67)** &t=-1.45 \\
        $\mu = -30$ &-29.52 (.57) &-29.12 (.95) &t=-.383 \\
        $\mu = -50$ &-50.86 (.52) &-50.37 (.66) &t=-.593 \\
      \hline
    \end{tabular}
  \end{center}
\end{table}

\subsection{Discussion}
Once again, we verify the \textit{biasing‑then‑debiasing} dynamic, providing direct support for the proposed model. Importantly, most generations ultimately align with the ground truth, further demonstrating that the model’s inherent planning ability remains intact but is initially interfered with. Moreover, we observe an immediate over‑correction following the initial starting bias. These results indicate that the model is aware of the correct response sequence from the outset and actively adjusts its biased generation in a dynamic, self‑adaptive manner.

\section{General Discussion}
\label{discussion}

We propose a Bayesian perspective on planning behavior in LLMs to account for the gap between their inherent planning ability and the compromised planning behavior observed at inference time. Through the lens of the evolving context during generation, we show how fundamental differences between how LLMs represent human language and self‑generated text provide a mechanism that drives shifts in internal planning, producing the appearance of short‑term planning. This framework not only offers a principled explanation for why planning becomes constrained at inference despite being supported by the training objective, but also predicts two additional signatures of planning behavior — \textit{planning convergence} and a \textit{biasing‑then‑debiasing} pattern — which we empirically validate in two experiments. Overall, our work deepens theoretical and empirical understanding of how LLMs plan ahead during inference.

Although compromised planning behavior has been reported in prior studies \citep{pal2023future, men2024unlocking}, such findings may reflect limitations in study design or measurement rather than a systematic underlying mechanism. Using two complementary experiments, our work consolidates the evidence for compromised planning at inference. Experiment 1 quantifies this compromise as reduced predictive performance and directly observes a short‑term planning pattern, along with increasing planning strength as in‑distribution context accumulates. Experiment 2 then demonstrates a biased beginning followed by an unbiased further continuation during generation. They reveal both potential fully planning ability and compromised actual planning behavior of LLMs, providing empirical supports to our proposed Bayesian account.

Sequence‑level planning enables LLMs to produce responses with higher sequence‑level likelihood. However, the initial biases revealed in this work suggest a mechanism that systematically distorts the sequence‑level distribution generated by LLMs, as illustrated in Figure \ref{generation_dynamics}. Techniques that recover or approximate the true sequence‑level distribution from model outputs may therefore help improve generation quality \citep[e.g.,][]{hu2026simulated}. An instructive example is self‑improvement, in which an LLM bootstraps its own outputs through iterative refinement \citep{huang2024self, song2024mind}. In a typical self‑improvement pipeline, the model first generates multiple candidate responses to a query and then refines them through self‑evaluation, such as voting \citep{huang2022large}, scoring \citep{liang2024sheep}, sampling‑based selection \citep{zelikman2022, song2024mind, wang2022self}, or other forms of feedback \citep{yuan2024self}. Substantial performance gains have been observed both in inference‑time refinement and in supervised fine‑tuning built on this paradigm \citep{SelfEvaluationImproves, piche2024selfevaluation}. The compromised planning behavior documented in our work may be the mechanism that drives these improvements.

\textbf{Limitations and Future Research.} A major limitation of this work concerns the generality of the results; specifically, whether free‑text generation exhibits the same planning patterns. Future work should extend the planning model to more unconstrained settings and evaluate whether the dynamics observed here persist when generation is less structured and more semantically open‑ended.







\printbibliography




\end{document}